\documentclass[11pt]{article}
\usepackage{graphicx} 

\usepackage{natbib}
\usepackage{multirow}
\usepackage{caption}
\usepackage{siunitx}
\usepackage{algorithm}
\usepackage{algpseudocode}
\usepackage{booktabs} 
\usepackage{fullpage}
\usepackage{authblk}
\usepackage{tikz}
\usepackage{mathtools}
\usepackage{amsmath, amssymb, amsfonts}
\usepackage{hyperref}
\usepackage{tikz}
\usepackage{bbding}
\usepackage{subcaption} 
\usepackage{gensymb}
\usepackage{fancyhdr}
\usepackage[top=2cm, headsep=1cm]{geometry}

\hypersetup{
    colorlinks=true,
    linkcolor=blue,
    filecolor=magenta,      
    urlcolor=cyan,
}

\usepackage{xcolor}

\title{Dynamic Targeting of Satellite Observations Using Supplemental Geostationary Satellite Data and Hierarchical Planning}

\author[1,2]{Akseli Kangaslahti\thanks{Correspondence to \texttt{steve.a.chien@jpl.nasa.gov}. Appears in the proceedings of the 2026 IEEE International Conference on Robotics \& Automation. \copyright 2026 California Institute of Technology. All rights reserved. Government sponsorship acknowledged. Personal use of this material is permitted. Permission from IEEE must be obtained for all other uses, in any current or future media, including reprinting/republishing this material for advertising or promotional purposes, creating new collective works, for resale or redistribution to servers or lists, or reuse of any copyrighted component of this work in other works.}}
\author[1,3]{Itai Zilberstein}
\author[1]{Alberto Candela} 
\author[1]{Steve Chien}

\affil[1]{Jet Propulsion Laboratory, California Institute of Technology, Pasadena, CA}
\affil[2]{Harvard University, Cambridge, MA}
\affil[3]{Department of Computer Science, Carnegie Mellon University, Pittsburgh, PA}

\begin{document}

\maketitle
\thispagestyle{fancy}

\pagestyle{fancy}
\fancyhf{}
\chead{\scriptsize Appears in Proc. of 2026 IEEE International Conference on Robotics and Automation, Vienna, Austria.} 
\renewcommand{\headrulewidth}{0pt}

\begin{abstract}
The Dynamic Targeting (DT) mission concept is an approach to satellite observation in which a lookahead sensor gathers information about the upcoming environment and uses this information to intelligently plan observations. Previous work has shown that DT has the potential to increase the science return across applications. However, DT mission concepts must address challenges, such as the limited spatial extent of onboard lookahead data and instrument mobility, data throughput, and onboard computation constraints. In this work, we show how the performance of DT systems can be improved by using supplementary data streamed from geostationary satellites that provide lookahead information up to 35 minutes ahead of time rather than the 1 minute latency from an onboard lookahead sensor. While there is a greater volume of geostationary data, the search space for observation planning explodes exponentially with the size of the horizon. To address this, we introduce a hierarchical planning approach in which the geostationary data is used to plan a long-term observation blueprint in polynomial time, then the onboard lookahead data is leveraged to refine that plan over short-term horizons. We compare the performance of our approach to that of traditional DT planners relying on onboard lookahead data across four different problem instances: three cloud avoidance variations and a storm hunting scenario. We show that our hierarchical planner outperforms the traditional DT planners by up to 41\% and examine the features of the scenarios that affect the performance of our approach. We demonstrate that incorporating geostationary satellite data is most effective for dynamic problem instances in which the targets of interest are sparsely distributed throughout the overflight.
\end{abstract}
 
\newpage
\setcounter{page}{1}

\section{Introduction} \label{introduction}

\begin{figure}[b]
    \centering
    \includegraphics[width=0.4\linewidth]{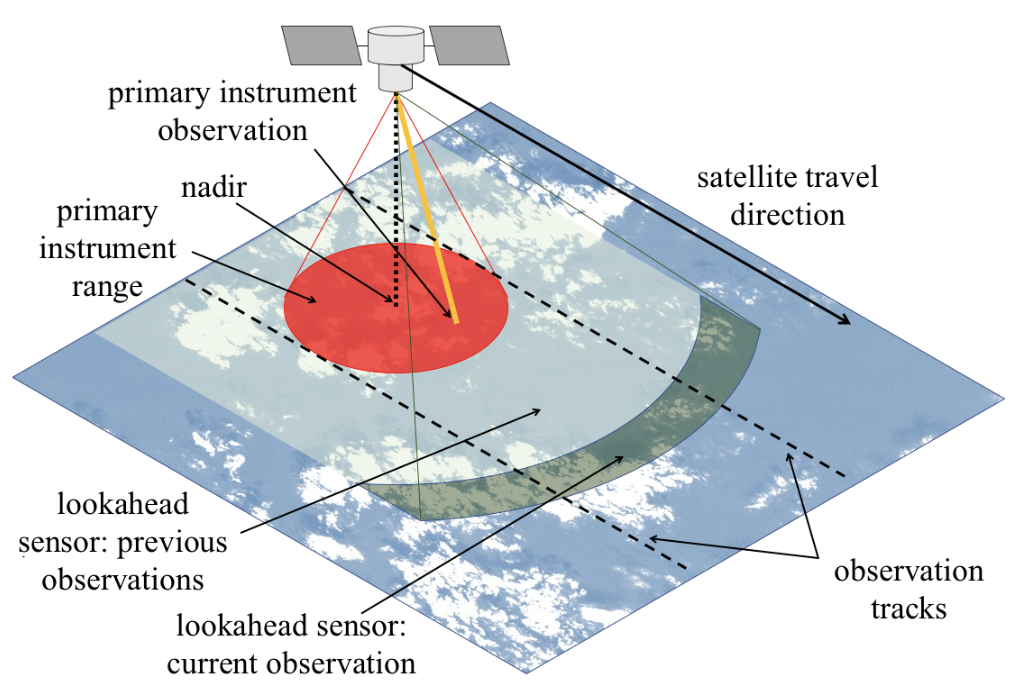}
    \caption{DT uses environmental information collected by a lookahead sensor to intelligently plan primary sensor observations. Figure adapted from \citep{dt-jais}.}
    \label{fig:dt-concept}
\end{figure}

Maximizing the scientific utility of a limited number of satellite observations is an ongoing challenge in remote sensing. Dynamic Targeting (DT) aims to address this problem with increased onboard autonomy. In the DT mission concept, a satellite gathers information about its upcoming environment with a cheap lookahead sensor, then uses this information to intelligently plan and execute highly valuable but expensive primary sensor observations. Figure \ref{fig:dt-concept} illustrates the DT mission concept.

In the DT planning problem, the onboard planner is tasked with deciding how to operate the primary sensor as to maximize the scientific utility of its observations while remaining within constraints. This is a sequential decision making problem over a discrete action space that recent work has tackled using search algorithms \citep{dt-planrob-2024}. However, there are several factors that make this an ongoing challenge, including the following:
\begin{itemize}
    \item \textit{Real-World Constraints}. Satellites are constrained by many factors that influence science return, including slewing (pointing) capability, energy consumption, and/or onboard computational and memory constraints.

    \item \textit{Real-Time Planning}. In this online planning application, the satellite is constantly moving at a fixed orbit velocity over the observation targets that are identified by the lookahead sensor. This means that the planner must generate plans quickly so that the satellite has time to slew to and observe planned observation targets before it flies past them.
    
    \item \textit{Limited Environment Information}. The spatial extent of the data from the lookahead sensor is minimal compared to the entire length of the orbit trajectory of the satellite. Thus, while the onboard lookahead sensor provides valuable information for short-term planning, optimally distributing a limited number of observations across the entire orbit trajectory without knowledge of the long-term environment remains challenging.  
\end{itemize}

Addressing these challenges is critical as several studies have shown DT's potential for improving observation efficiency across many applications \citep{dt-jais, candela-pbl-igarss2024}. Prior work has shown that DT is especially promising for studies of rare, dynamic events \citep{Zilberstein25:Real}. Possible Earth science applications include cloud avoidance, storm hunting, harmful algal bloom monitoring, wildfire observations, volcanic plume studies, and more. Even outside of Earth science, DT can potentially be used to observe plumes that erupt on the surfaces of comets and icy moons like Europa and Enceladus \citep{Brown_2019}. Since DT has already shown promise to improve science return in many of these applications, we expect it to become common among future missions \citep{dt-jais}. 

In this work, we explore the use of geostationary satellite data to address the problem of having only short-term environmental information. Geostationary satellites remain above and constantly observe the same area of the Earth with cadence often on the order of tens of minutes and coverage that can span continents. Recent research has introduced the idea of using geostationary satellite data along the orbit trajectory of a satellite as a sole ``virtual lookahead" with experiments on a single problem instance that show how latency in transmitting geostationary satellite data to the primary satellite affects performance \citep{kacker-iwpss-2025}. We build off of these studies in this work by studying how incorporating geostationary satellite data along the orbit trajectory of a traditional DT satellite as an ``extended lookahead" that supplements the data collected by the onboard lookahead sensor can improve on the performance of the traditional DT mission concept that only uses onboard lookahead sensor data. First, we introduce a hierarchical planning framework for DT to address the increased problem complexity induced by the extension of the lookahead data range. Then, we conduct simulation studies of four problem instances in which we directly compare the performance of planners that use only onboard lookahead data to that of a planner that leverages both onboard lookahead data and supplemental geostationary satellite data. We analyze our results as well as the properties of the problem instances that affect the performance of each approach to help understand which DT applications can benefit most from the integration of geostationary satellite data.

\section{Related Work} \label{related-work}
The core DT concept of an agent planning and executing intelligent actions based on its sensor data is quite common in robotics, even outside of aerospace and satellite technologies \citep{vision, lidar}. As in the online Informative Path Planning problem \citep{schmid2020efficient}, the goal is to adaptively plan trajectories that maximize a utility function as new information emerges. However, traditional DT satellites are constrained by the satellite orbit trajectory and a fixed environment discovery process via the lookahead sensor, and are thus more focused on observation utility maximization within a locally known portion of the environment rather than information gathering. Strategies analogous to the ``extended lookahead" that we use in this work can also be leveraged in other robotic domains. For example, prior research has shown how drones can be used in search and rescue applications to create a map of a large environment that humanoid robots can then use to supplement information from their own sensors during rescue missions \citep{drones}. 

Within DT, early work focused on simulation studies that proved the DT concept and used simpler planning algorithms intended for storm hunting \citep{Swope2024:Storm} and cloud avoidance \citep{dt-jais} applications while considering fewer constraints and less complex utility models. More recently, realistic slewing constraints and adaptive utility models have been introduced to DT simulation studies, introducing a sequential element to the DT planning problem that motivates the use of search algorithms \citep{dt-icra-2024, dt-astra-2023}. A subsequent study performed a more thorough analysis of this search problem, drawing insights regarding tradeoffs between planning time vs. execution time as well as exploration vs. exploitation in planning algorithms \citep{dt-planrob-2024}. In this work, we use a previously proposed slew model \citep{dt-icra-2024} and test our algorithms on variations of the cloud avoidance and storm hunting applications. Learned approaches have also been proposed for DT observation planning \citep{Breitfeld25:Learning}. Finally, beyond simulation studies, DT has also been deployed on several real-world flights \citep{Zilberstein24:Demonstrating, Zilberstein25:Rapid, Chien24:Leveraging, Chien25:Dynamic}.

Additionally, DT is only a subset of a larger effort to improve the efficiency of satellite observation collection. Outside of DT, operational work includes screening cloudy observations out of onboard data storage for the Airborne Visible-Infrared Imaging Spectrometer - Next Generation (AVIRIS-NG) instrument \citep{screen}. Another example is the Thermal and Near Infrared Sensor for Carbon Observation Fourier-Transform Spectrometer-2 (TANSO-FTS-2) instrument, which used intelligent targeting to increase the number of clear-sky observations by a factor of 1.8 \citep{tanso-fts-2}. The FORMOSA-2 satellite also used a form of intelligent targeting that leveraged weather conditions and forecasts to generate an observation schedule that could be adjusted at execution time \citep{weather}. Hypothetical simulation studies include a feasibility study for an onboard lookahead camera that can help perform cloud avoidance \citep{feasibility} and algorithms for intelligent observation planning based on previous observations that showed promise on offline datasets \citep{ase, ase_improved}. Recent research has also shown, in simulation, how geostationary satellite data along the future orbit trajectory of a primary satellite can be leveraged as a sole ``virtual lookahead" to plan to avoid cloudy observations when fulfilling tasking requests \citep{kacker-iwpss-2025}. This study analyzed how varying latency in transmitting geostationary satellite data to the primary satellite affects performance the planner. In our work, we build on this research by incorporating geostationary satellite data into the traditional DT mission concept as an ``extended lookahead" that supplements the onboard lookahead sensor data and testing on four different problem instances. We focus our analysis more on the quantitative performance effect of incorporating the ``extended lookahead" compared to using only onboard lookahead sensor data, and on understanding which problem instances this strategy is best suited for.

\section{Methods} \label{methods}
\subsection{Simulation Studies} \label{simulation-datasets}
We use two different datasets to simulate onboard lookahead data in our experiments:
\begin{enumerate}
    \item \textit{MODIS Cloud Mask}. The Moderate Resolution Imaging Spectrometer (MODIS) is an instrument onboard the Aqua and Terra satellites of NASA that provides visible spectrum data. In this work, we use a preprocessed binary cloud mask with a spatial resolution of about 1 km/pixel that NASA derives from MODIS data collected onboard the Aqua satellite and publishes online \citep{modis}. We download 19 35-minute flights of this data over North and South America, each from a different day of December 2024. We use this dataset for variations on the cloud avoidance DT application, which we describe in Section \ref{utility-models}.
    \item \textit{IMERG}. The Integrated Multi-satellitE Retrievals for GPM (IMERG) dataset is a global precipitation measurement product derived from observations made by satellites in the Global Precipitation Measurement (GPM) satellite constellation. We use the half-hourly ``Final Run" product \citep{imerg}, which provides continuous mm/hr precipitation values at a spatial resolution of about 10 km/pixel and is the recommended product for research purposes. We download 21 global data arrays, each from a random time on different days of January 2018. This dataset serves as a proxy for storms, and we use it for the storm hunting DT application, which we describe in Section \ref{utility-models}.
\end{enumerate}

\begin{figure}
    \centering
    \includegraphics[width=0.5\linewidth]{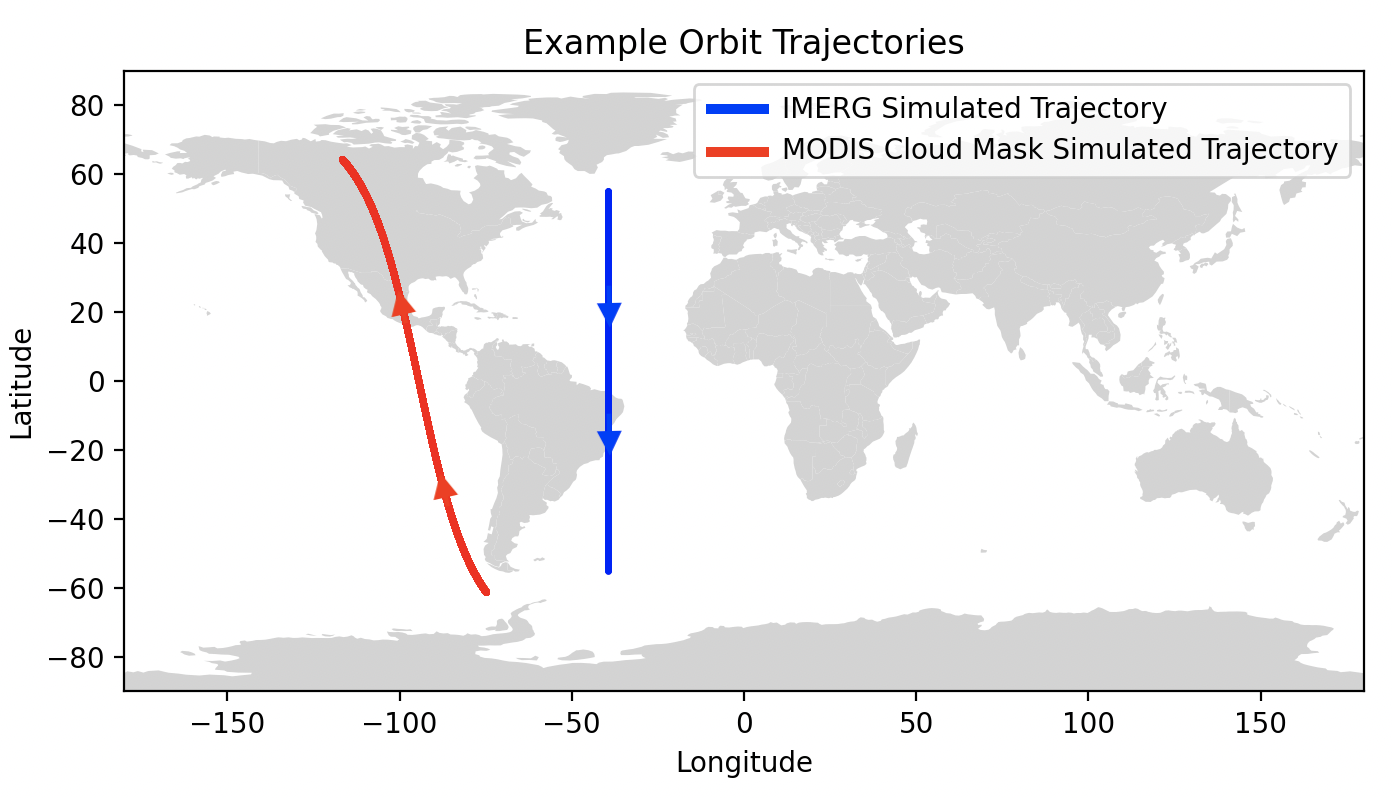}
    \caption{Example simulated orbit trajectories for each type of simulation data. The red line represents an orbit trajectory for the MODIS Cloud Mask data, while the blue line represents an orbit trajectory for the IMERG data.}
    \label{fig:trajectories}
\end{figure}

When simulating a flyover, we use one of the two datasets described above as the simulated onboard lookahead data, with each pixel in the dataset representing a possible observation target. The MODIS Cloud Mask data comes from the MODIS sensor aboard the Aqua satellite of NASA, so the simulated orbit trajectories for simulations that use this data are simply the orbit trajectories that the Aqua satellite took when collecting the data. In contrast, the IMERG dataset is a global product that is derived from multiple satellites. To simulate an orbit trajectory for storm hunting flyovers, we first randomly select a longitude between $55\degree$ W and $55\degree$ E. The selected longitude line is different for each of the 21 days of data. We then trace this longitude line from $55.05\degree$ N to $55.05\degree$ S with a constant orbit velocity of about 6 km/second and use the resulting trajectory as the simulated orbit trajectory. Although this is not a realistic satellite trajectory, similar simplified trajectories have been used in prior DT studies \citep{Swope2024:Storm}. The simulated onboard lookahead data for the flyover is the data from the IMERG product along that trajectory, captured with a swath of about 270 km. We can simulate 19 35-minute flyovers using our MODIS Cloud Mask data and 21 30-minute flyovers using our IMERG data, each of which has a different simulated flight trajectory and corresponding simulated onboard lookahead data. Figure \ref{fig:trajectories} shows an example orbit trajectory for each type of data.

Different types of planners use different modes of data to make decisions. Planners that use only onboard lookahead data are given access to simulated lookahead data along the simulated trajectory about 500 km ahead of the position of the satellite throughout the flyover, since our simulated orbit altitude is 500 km and the lookahead sensor is assumed to be pointed in front of the satellite at a $45\degree$ angle. Planners that also leverage geostationary satellite data get access to additional cloud mask or precipitation data that effectively extends the lookahead distance but may not perfectly predict the onboard lookahead data, which is assumed to be the ground truth. The MODIS Cloud Mask data is supplemented by a Clear Sky Mask data product from the Advanced Baseline Imager (ABI) of the Geostationary Operational Environmental Satellite (GOES) East and West satellites, which cover the North and South America continents \citep{goes}. This product has a spatial resolution of 2 km/pixel and a cadence of 10 minutes. The IMERG data is supplemented by the Multi-Sensor Precipitation Estimate product provided by EUMETSAT \citep{meteosat}, which aggregates data from the series of Meteosat geostationary satellites to produce a product with a spatial resolution of about 4 km/pixel and a cadence of 15 minutes that spans most of the Atlantic Ocean and Africa, with a longitude range of $67.5\degree$ W to $67.5\degree$ E and latitude range of $67.5\degree$ S to $67.5\degree$ N. We download these geostationary data products in the timeframe that matches that of the simulated onboard lookahead data and query them along simulated orbit trajectories using a nearest-neighbor algorithm so that planners that leverage the supplemental geostationary data can have an ``extended lookahead" that spans their entire simulated orbit trajectory.

\subsection{Utility Models} \label{utility-models}
We use four different utility models in our experiments. Each utility model is a function of the onboard lookahead data that indicates the scientific utility of each observation target. We use powers of 10 consistently across our utility models. The utility models are explained below.
\begin{itemize}
    \item \textit{Cloud Avoidance (CA)}. The goal of this classic DT application is to collect visible spectrum observations in clear skies and avoid observations blocked by clouds. The CA utility model is only used in simulations with the MODIS Cloud Mask data. In CA, an observation of a pixel marked as cloudy in the MODIS Cloud Mask results in 1 utility, while an observation of a pixel marked as clear sky in the MODIS Cloud Mask results in 10 utility. This is intended to encourage clear sky observations over those that are blocked by clouds.
    
    \item \textit{Cloud Avoidance with Population Density (CAPD)}. This utility model is only used in simulations with the MODIS Cloud Mask data. In CAPD, an observation of a pixel marked as cloudy in the MODIS Cloud Mask results in 1 utility. For clear sky observations, the utility $u$ is calculated as follows:
    \begin{equation} \label{pop-utility}
        u = 10^{\frac{p}{10000} + 1}
    \end{equation}
    In Equation \ref{pop-utility}, $p$ refers to the population density, in people per square kilometer, in the area covered by the observation target. We use a global population density dataset based on census data from 2000 provided by NASA Earth Observations (NEO) \citep{population} to find $p$. This utility model is intended to especially encourage clear-sky observations of densely populated areas, which are where tasking requests are often targeted.

    \item \textit{Cloud Avoidance with Random Targets (CART)}. This utility model is only used in simulations with the MODIS Cloud Mask data. In CART, there are 20 target areas shaped like a square with a random side length between 20 and 100 km that are randomly distributed throughout the simulated flyover. An observation of any pixel marked as cloudy in the MODIS Cloud Mask data results in 1 utility. An observation of a pixel that is marked as clear sky in the MODIS Cloud Mask data and is located inside a target area results in 100 utility. An observation of any other pixel that is marked as clear sky in the MODIS Cloud Mask data results in 10 utility. This is intended to roughly represent a generalization of missions where there are some static targets that are known ahead of time and that we are interested in observing during the flyover. Example targets could include natural anomalies like volcanoes, military applications, or commercial tasking requests.

    \item \textit{Storm Hunting (SH)}. Storm hunting is another classic DT application where the goal is to observe dyanmic storm events. The SH utility model is only used in simulations with the IMERG precipitation data. In SH, the utility $u$ of a pixel is calculated as follows:
    \begin{equation}\label{sh-utility}
        u = 100^{\frac{r}{r_{max}}}
    \end{equation}
    In Equation \ref{sh-utility}, $r$ indicates the precipitation rate in the pixel in mm/hr and $r_{max}$ indicates the maximum precipitation rate across all the IMERG data from our 21 simulated flyovers. This is intended to encourage observations of targets with high precipitation rates, which we use as a proxy for storms.
\end{itemize}

\begin{figure}
    \centering
    \includegraphics[width=0.5\linewidth]{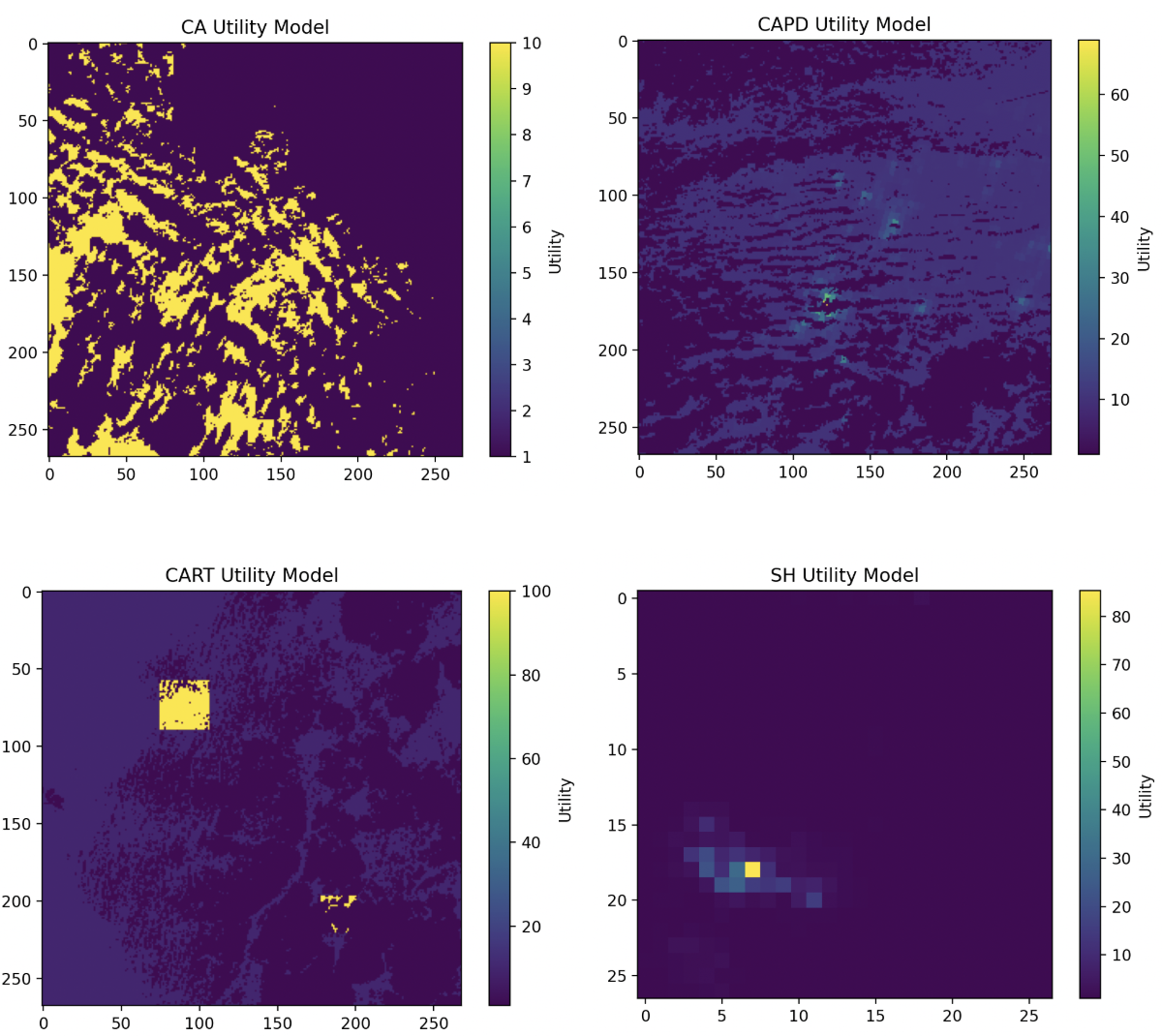}
    \caption{Examples of the CA (top left), CAPD (top right), CART (bottom left), and SH (bottom right) utility models applied to parts of the MODIS Cloud Mask and IMERG datasets. Axes denote pixels, which have 1 km and 10 km spatial resolutions for MODIS-based and IMERG-based utility models, respectively. All images cover an area of about 260 km $\times$ 260 km. Note that the spatial resolution of the MODIS Cloud Mask dataset is finer than that of the IMERG dataset, which is why the SH utility model example appears coarser.}
    \label{fig:util-models-examples}
\end{figure}

Figure \ref{fig:util-models-examples} shows an example snapshot of each utility model.

\subsection{Problem Statement} \label{problem-statement}
\begin{figure}
    \centering
    \includegraphics[width=0.25\linewidth, angle = 270]{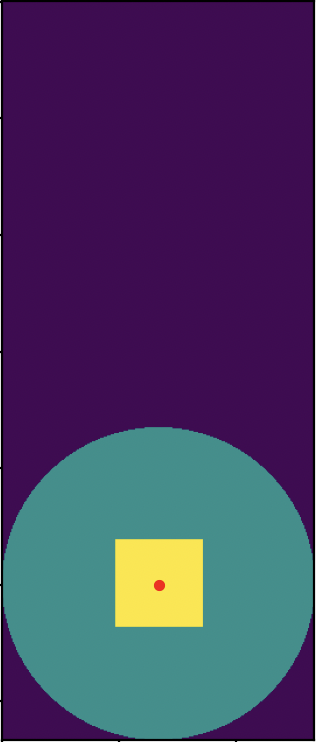}
    \caption{A visualization of the primary sensor slewing capability. The right edge of the figure represents where the lookahead sensor is sensing new data. The satellite is traveling towards the right edge of the figure. The primary sensor is current pointed at nadir, which is indicated by the red dot. The yellow area shows the reachable set of observation targets that the primary sensor can be slewed to within one cycle (4 seconds) of slewing time. The turquoise blue area shows the entire primary sensor range, which is determined by the primary sensor's $15\degree$ maximum off-nadir. The purple area represents observation targets that have been sensed by the lookahead sensor but could not be reached during this cycle regardless of the initial position of the primary sensor.}
    \label{fig:slew-range}
\end{figure}

The basic components of the DT planning problem are:
\begin{itemize}
    \item $F$: The simulated flight trajectory of the satellite.
    \item $D$: The data that $F$ passes over, which is sensed over time by the lookahead sensor of the satellite.
    \item $S$: The set of possible states that the satellite can be in.
    \item $A$: The set of possible actions that the satellite can take.
    \item $C$: The set of constraints on the satellite.
    \item $U$: The utility model.
    \item $T$: The transition function.
\end{itemize}
$F$ is discretized into 4 second cycles, which are discrete partitions of time during which exactly one observation can be taken. For example, the IMERG dataset flyovers each last 30 minutes (1,800 seconds), so they span 450 cycles. Each cycle is associated with a state, which specifies:
\begin{itemize}
    \item $(x, y)$, the current primary sensor position as an index of the data array $D$.
    \item $o$, the number of observations that have been collected during the flyover so far.
    \item $t$, the number of cycles that have passed so far.
\end{itemize}
During cycle $c$, the planner has access to $D_c \subset D$, the portion of the data that the onboard lookahead sensor has sensed by cycle $c$, which it can use to decide the next action(s). If the planner also integrates the supplemental geostationary satellite data, it can leverage information about the rest of the orbit trajectory, but the geostationary satellite data is not necessarily a perfect match with $D$. The satellite executes a single action during each cycle. An action $a \in A$ specifies the following:
\begin{itemize}
    \item $(x^\prime, y^\prime)$, the new primary sensor position.
    \item $p$, a boolean value that specifies whether the primary sensor will be turned on for an observation at $(x^\prime, y^\prime)$.
\end{itemize}
If $p$ is True, an observation will be collected, and utility will be awarded as specified by $U$. Regardless of the value of $p$, $T$ is used to transition to the next state. In the subsequent state, the satellite moves further along $F$ and has access to a slightly greater subset of $D$. However, the actions that the satellite takes must not violate the constraints $C$:
\begin{itemize}
    \item \textit{Observation Volume}. During all simulated flyovers, the satellite is restricted to 100 total observations. This represents onboard storage and energy constraints. Since all simulated flyovers last 30-35 minutes, the primary sensor can only be turned on for an observation during about 20\% of cycles.
    \item \textit{Slewing}. The primary sensor of the satellite can be pointed along the roll and pitch axes, but slewing maneuvers take time. The sensor must also be at rest for an observation to be collected. As in prior work \citep{dt-icra-2024, dt-planrob-2024}, we model slewing constraints with a constant angular acceleration/deceleration of $\alpha = 1.08\degree$/s and a maximum angular velocity of $\omega = 5.40\degree$/s, which are realistic values for a small satellite that uses control moment gyroscopes (CMGs) for slewing \citep{LAPPAS}. These values apply to both slewing axes. Figure \ref{fig:slew-range} shows an example of how far the primary sensor can be slewed with one cycle (4 seconds) of slewing time. This constraint is what makes the DT planning problem a sequential problem, since slewing during cycle $c$ affects the actions that are available during cycle $c+1$.
\end{itemize}

The goal of a DT planner is to, over the course of $F$, plan and execute an action sequence $\pi = \{a_1, a_2, a_3,...,a_n\}$ (where $n$ is the total number of cycles in $F$ and $a_i \in A$ $\forall i$) that does not violate $C$ and maximizes the total utility of the observations, as determined by $U$.

\subsection{Planning Algorithms} \label{planning-algorithms}
We experiment with baseline algorithms and hierarchical planners. Baseline algorithms typically plan and execute one action at a time and are mainly intended to put the performance of hierarchical planners in perspective. Hierarchical planners use a high-level planner to distribute observations throughout the simulated flyover, then use search algorithms to repeatedly produce and execute short-term action plans that follow the pre-planned observation distribution.
\subsubsection{Baseline Algorithms} \label{baseline-planners}
\begin{itemize}
    \item \textit{Nadir Only (NO)}. This algorithm only makes observations at nadir (the point directly below the satellite) and is indifferent to any lookahead data. The 100 observations are evenly distributed throughout the simulated flyover, so that an at-nadir observation is collected roughly once every 5 cycles. This is representative of most planners aboard current Earth science missions.
    \item \textit{Greedy (G)}. This algorithm greedily expends all 100 observations during the first 100 cycles. During each of these first 100 cycles, the Greedy algorithm takes the observation with the highest immediate utility that is available in the reachable set of observation targets.
    \item \textit{Greedy Historical (GH)}. During each cycle, this algorithm finds $u_{max}$, the highest utility observation that is available in the reachable set of observation targets. It incrementally builds an array of $u_{max}$ values, one for each cycle. The Greedy Historical algorithm only makes an observation of the target with $u_{max}$ utility if $u_{max}$ is greater than the mean of the recorded array so far. The idea of this algorithm is to save observations for targets that, according to historical data, are ``above average" observations.
    \item \textit{Upper Bound (UB)}. This algorithm is only for performance comparison and does not produce a feasible observation plan. First, it omnisciently constructs an array containing the highest utility observation target within the entire primary sensor range during every cycle in the simulated flyover. It then returns the sum of the highest 100 values in this array. Since this ignores slewing constraints, the Upper Bound algorithm is not necessarily tight.
\end{itemize}

\subsubsection{Hierarchical Planners} \label{hierarchical-planners}
All hierarchical planners follow the same structure, which is outlined in Algorithm \ref{pseudocode}.

\begin{algorithm}
    \caption{Planning procedure used by all hierarchical planning algorithms.}\label{pseudocode}
    \begin{algorithmic}
        \While {simulation.notOver()}
            \If {newGeoDataAvailable(simulation.cycle)}
                \State data $\gets$ getUpdatedGeostationaryData()
                \State distribution $\gets$ distributeObservations(data)
            \EndIf
            \State plan $\gets$ generatePlan(distribution[simulation.cycle])
            \State executePlan(plan)
        \EndWhile
    \end{algorithmic}
\end{algorithm}

During each iteration of the algorithm, we first check if the geostationary data that we have can be updated. For both the MODIS Cloud Mask and the IMERG simulations, this update happens once at the beginning of the simulation to initialize the geostationary satellite data. In the MODIS Cloud Mask simulations, we update the geostationary satellite data from GOES three more times throughout the 35-minute simulated flyover, since the GOES data has a cadence of 10 minutes. Unlike the whisk-broom data collected by Aqua that we use in MODIS Cloud Mask simulations, the IMERG data that we use for onboard lookahead data in storm hunting simulations is a snapshot that is not collected in real time throughout the simulated trajectory. Thus, we initialize the geostationary satellite data from the Meteosat series with an acquisition from the beginning of the half-hourly IMERG collection period and do not update it further during the simulated flyover.

The \textit{generatePlan} function from Algorithm \ref{pseudocode} is responsible for generating short-term plans for each of the flyovers, which we divide into 10-cycle partitions. It uses a Beam Search algorithm with a search depth of 10 and a beam width of 3, similar to that used in prior work \citep{dt-icra-2024, dt-planrob-2024}. The limited beam width is a simplified way of enforcing onboard computational and memory constraints and real-time planning constraints. Since it plans 10 actions at once, it takes advantage of a greater portion of the simulated onboard lookahead data than the baseline planning algorithms. It takes an integer input that specifies how many observations it can make during each 10-action plan that it generates, which is specified by the observation distribution. The search algorithm has the additional advantage of being able to spend multiple cycles slewing in order to reach faraway targets. For example, if an observation is made during cycle $c$, the search algorithm can plan to slew to and observe a faraway target during cycle $c+2$ or later. This option has been described in more detail in prior work \citep{dt-icra-2024} and is always one of the three children of each node that is expanded in the Beam Search. The remaining two children are the two highest utility observations that can be made during the next cycle. The search returns the 10-cycle observation plan with the greatest total utility that it finds.

The \textit{distributeObservations} function in Algorithm \ref{pseudocode} is responsible for distributing the 100 across the 10-cycle partitions of the simulated flyover. To make this decision, it can consider the geostationary satellite data. It can also replan its distribution when updated geostationary satellite data is received. This is the only time when the geostationary satellite data is used for planning. We experiment with three algorithms for distributing observations across the 10-cycle partitions of the simulated flight trajectory, which we describe below.

\begin{itemize}
    \item \textit{Uniform (U)}. This algorithm distributes the 100 observations uniformly across the 10-cycle partitions. This works out to about 2 observations per partition. This approach ignores geostationary satellite data and is representative of a DT planner that only uses onboard lookahead data.
    \item \textit{Preinformed (P)}. This algorithm distributes observations proportionally to any static targets that can be determined in advance without using lookahead data. Targets are only known in advance for the CAPD utility model, where the population density along the simulated orbit trajectory is known, and the CART utility model, where the locations of the special 100-utility targets are known in advance. As such, we do not use this algorithm for the CA or SH utility models. In the Preinformed algorithm, we sum the potential utilities of all the known target areas in each of the 10-cycle partitions of the simulated orbit trajectory, assuming that none of them are blocked by cloud cover. We then distribute observations across partitions proportionally to utility sums within partitions. This approach ignores geostationary satellite data and is representative of a more intelligent DT planner that still only uses onboard lookahead data.
    \item \textit{Geostationary Satellite Data-Informed (GSDI)}. This algorithm uses the supplemental geostationary satellite data to intelligently distribute observations proportionally to the anticipated utility within partitions. It plugs the geostationary satellite data along the entire orbit trajectory into the utility model to produce a matrix of anticipated utility for the whole simulated flyover. Similarly to the Preinformed algorithm, it then sums the anticipated utility within each partition and distributes observations across partitions proportionally to these sums. This is the only approach that leverages the geostationary satellite data. One weakness of this approach under our experiment design is that the geostationary satellite data may differ from the simulated onboard lookahead data, which is used as the ground truth in the utility models.
\end{itemize}
\section{Experiments and Results} \label{results}
First, we study the predictive power of our geostationary satellite data relative to our simulated onboard lookahead data. We recognize that the strategy of using the simulated onboard lookahead data as the ground truth is flawed because the MODIS Cloud Mask and IMERG data are still prone to error. However, our results show that DT planners that use only onboard lookahead data can be surpassed in performance by planners that also use supplemental geostationary satellite data, even when the onboard data is given the advantage of being the ground truth.

\begin{figure}
    \centering
    \includegraphics[width=0.4\linewidth]{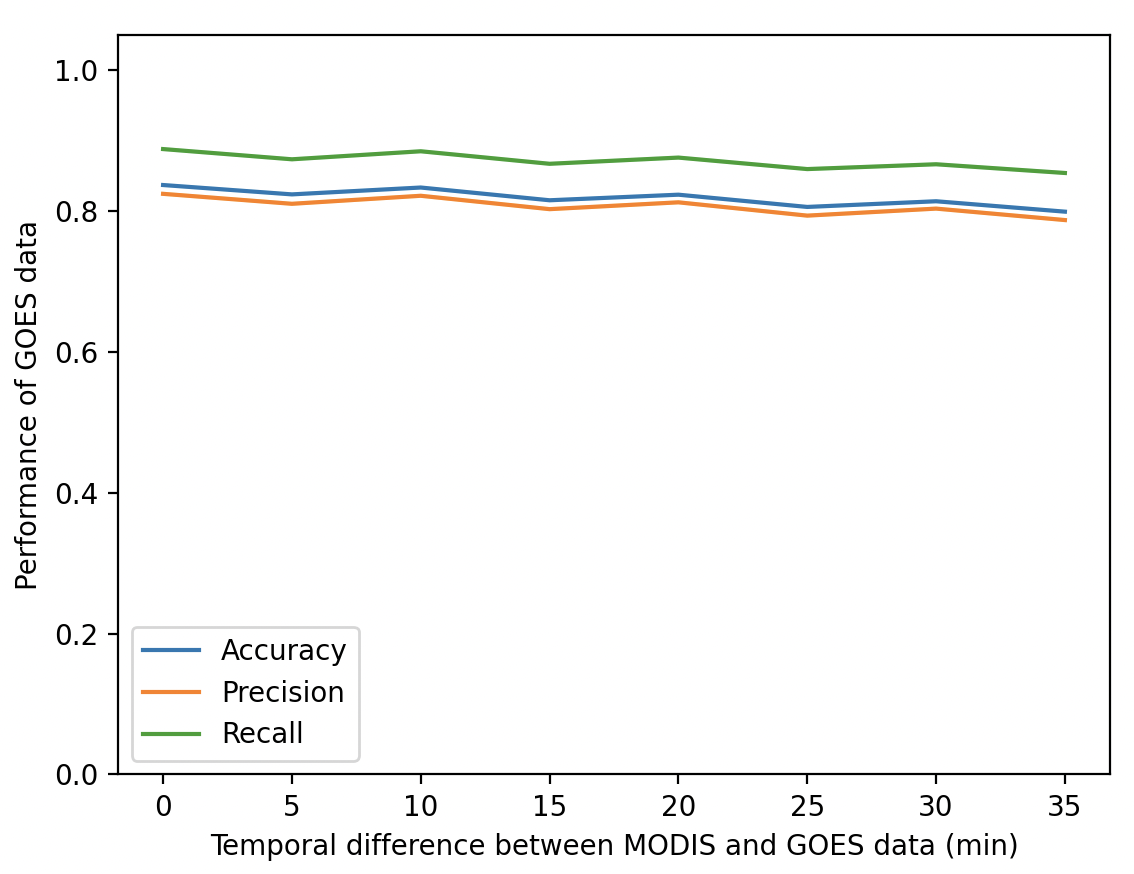}
    \caption{Average predictive performance of the GOES Clear Sky product, with the MODIS Cloud Mask as ground truth.}
    \label{fig:modisvgoes}
\end{figure}

For the MODIS Cloud Mask and GOES Clear Sky mask data, we calculate the average accuracy, precision, and recall of the GOES data across the 19 simulation datasets, using the MODIS data as ground truth. We test with 0-35 minutes of latency in the GOES data to match the simulated flyover length. Figure \ref{fig:modisvgoes} shows the results of this experiment. In this case, a ``positive" is a pixel marked as cloudy in the GOES data and a ``negative" is a pixel marked as clear sky in the GOES data.

The predictive power of the GOES data, relative to the MODIS data, remains mostly constant when varying the latency from 0-35 minutes. However, the accuracy is only about 80\%, even with 0 latency. The recall is about 6\% higher than the precision, meaning the GOES data has fewer false negatives than false positives.

For the IMERG and Meteosat data, we cannot make relevant claims about the relationship between the predictive power of the Meteosat data and temporal latency because we simulate flyovers over a static snapshot of IMERG data and we do not update the Meteosat data throughout any of the simulated flyovers. Instead, we calculate that the mean absolute error of the single Meteosat dataset used per IMERG simulation is 0.190 mm/hr. Among pixels marked as having nonzero precipitation in either product, which represents only 16.8\% of pixels, the mean absolute error of the Meteosat dataset is 1.13 mm/hr. One advantage of our Geostationary Satellite Data-Informed algorithm is that the geostationary satellite data does not have to provide perfect pixel-level predictions as long as it captures the general patterns well enough to make intelligent high-level decisions about how many observations to allocate to each partition. Although we do not address uncertainty in the geostationary satellite data, our results show that it can still outperform traditional DT planners. 

Finally, for each planning algorithm, we run all 19 simulations with the MODIS Cloud Mask data in combination with the CA, CAPD, and CART utility models and all 21 simulations with the IMERG data paired with the SH utility model. The aggregate utilities accrued by each planning algorithm are shown in Table \ref{final-results} as a percentage of the UB utility. Recall that the Preinformed algorithm (P) is only applicable to the CAPD and CART utility models.

\begin{table}
\caption{Aggregate Utility Accrued by Each Planning Algorithm Across Different Simulation Scenarios.}
\label{final-results}
\centering
\begin{tabular}{@{} l cccc @{}}
\toprule
\textbf{Algorithm} & 
\multicolumn{3}{c}{\textbf{MODIS}} & 
\multicolumn{1}{c}{\textbf{IMERG}} \\
\cmidrule(lr){2-4} \cmidrule(l){5-5}
& 
\multicolumn{1}{c}{CA$^1$} & 
\multicolumn{1}{c}{CAPD$^2$} & 
\multicolumn{1}{c}{CART$^3$} & 
\multicolumn{1}{c}{SH$^4$} \\
\midrule

NO         & 41.9\%  &  9.6\%  &  4.8\% & 24.4\% \\
G          & 63.8\%  & 14.8\%  & 11.9\% & 27.1\% \\
GH         & 88.2\%  & 20.4\%  & 25.0\% & 24.7\% \\
U          & 97.5\%  & 49.9\%  & 44.7\% & 47.5\% \\
P          & --      & 52.0\%  & \textbf{64.9\%} & -- \\
GSDI       & \textbf{99.9\%} & \textbf{72.8\%} & 63.3\% & \textbf{67.2\%} \\

UB         & 100.0\%  & 100.0\%  & 100.0\% & 100.0\% \\ 
\bottomrule
\end{tabular}

\vspace{0.3cm}
\begin{minipage}{\textwidth}
\footnotesize
\textsuperscript{1}CA = Cloud Avoidance \\
\textsuperscript{2}CAPD = Cloud Avoidance with Population Density \\
\textsuperscript{3}CART = Cloud Avoidance with Random Targets \\
\textsuperscript{4}SH = Storm Hunting

\end{minipage}
\end{table}

First, we compare the performance of the hierarchical planners to that of the baseline algorithms. Across all four problem instances, the Nadir Only algorithm is outperformed by all other DT algorithms, with the best DT algorithms often performing orders of magnitude better than the Nadir Only algorithm. This highlights the potential that the DT mission concept has to improve satellite observation efficiency. Furthermore, the best hierarchical search-based planner improves on the performance of the best greedy algorithms by an average factor of 2.44, which validates the hierarchical planning structure. Although the best performing hierarchical planner nearly matches the Upper Bound under the CA utility model, they fail to capture the remaining 27.2-35.1\% of the Upper Bound utility under the other utility models. It is unclear how much of this gap can be attributed to the fact that the Upper Bound algorithm is not tight. However, the hierarchical planners are reaching high percentages of the Upper Bound utility compared to other DT work \citep{dt-planrob-2024, dt-icra-2024}.

Next, we compare the hierarchical planners that use only the simulated onboard lookahead data (Uniform and Preinformed) to the hierarchical planner that also incorporates the supplemental geostationary satellite data (Geostationary Satellite Data-Informed). This is perhaps the most important comparison, as it isolates the effect of using the geostationary satellite data. The Geostationary Satellite Data-Informed algorithm strongly outperforms the better of the Uniform and Preinformed planners under the CAPD and SH utility models by a factor of 1.40 and 1.41, respectively. Under the CA and CART utility models, it roughly matches the performance of the better of the Uniform and Preinformed planners. We believe that this is related to the distribution of high-utility targets within the different utility models. The mean and maximum observation utilities are about 4.24 and 10, respectively, in the CA utility model and 4.90 and 100, respectively, in the CART utility model. In contrast, the mean and maximum observation utility are about 4.26 and 1477, respectively, in the CAPD utility model and 1.02 and 100, respectively, in the SH utility model. This indicates that in the CA and CART utility models, targets with utility near the maximum observation utility come by more frequently, so the long-term observation distribution does not matter as much because the search algorithms can often reactively find high-utility observations. However, in the CAPD and SH utility models, there are sparse clusters of (relatively) extremely high utility targets in the flyovers, during which planners need to allocate as many observations as possible. Thus, the better-informed Geostationary Satellite Data-Informed hierarchical algorithm has the greatest advantage under such utility distributions. This observation is consistent with the sparse small-cluster nature of cities and storms, and it aligns with prior analysis of cloud avoidance vs. storm hunting in DT \citep{dt-planrob-2024}. Additionally, although the Geostationary Satellite Data-Informed algorithm always has the advantage of having additional information about the long-term environment, this advantage is especially strong in entirely dynamic applications like SH where traditional DT planners have zero information about the long-term environment.

\section{Conclusion} \label{conclusion}
DT has shown promise to increase science return of Earth science missions, especially when observing rare, dynamic events \citep{Zilberstein25:Real}. However, optimizing DT planning algorithms remains an ongoing challenge. To address the problem of only having access to short-term environment information during planning, we examine through a series of simulation studies how DT planners can incorporate geostationary satellite data to supplement onboard lookahead data. We introduce a new hierarchical planning structure for DT that can intelligently handle the increased problem complexity induced by the extended lookahead range and experiment with four different utility models. Our results show that incorporating geostationary satellite data into DT planners is most effective for dynamic applications where relatively high-utility observation targets are sparsely distributed throughout flyovers.

In future work, we aim to evaluate more planning strategies for DT with supplemental geostationary satellite data. For example, algorithms that directly use the extended lookahead to make deeper plans that go past the onboard lookahead data range may be effective. We also aim to design other algorithms that consider the uncertainty of the geostationary satellite data during planning. Another planning strategy could be to make a tentative observation distribution using geostationary satellite data that can adjusted in real time based on the data that the onboard lookahead sensor collects. We also plan to consider a ``just in time" commanding strategy in which we process the geostationary satellite data on the ground rather than onboard the spacecraft then command the spacecraft from the ground as to maximize observation utilities. Additionally, we are interested in building on multi-agent DT formulations with multiple satellites carrying primary instruments that collaboratively make observations and convey information about the environment to each other \citep{madt1, madt2, Zilberstein26:Large}. Finally, we are working to expand on efforts to benchmark DT algorithms on more realistic flight processors \citep{fame-spaceops-2025} and to get DT systems deployed on more missions. Nevertheless, we have shown in this work how geostationary satellite data can be incorporated into DT planners to supplement onboard lookahead data and empirically analyzed the problem instances for which this strategy is most promising. 

\section*{Acknowledgments}
    This research was carried out at the Jet Propulsion Laboratory, California Institute of Technology, under a contract with the National Aeronautics and Space Administration (80NM0018D0004). This work was supported by the Earth Science and Technology Office (ESTO), NASA.  
\bibliography{arxiv_references}

\end{document}